%% file: acl_latex.tex
\pdfoutput=1

\documentclass[11pt]{article}

\usepackage{acl}

\usepackage{times}
\usepackage{latexsym}
\usepackage{amssymb}
\usepackage{float}

\usepackage[T1]{fontenc}

\usepackage[utf8]{inputenc}

\usepackage{microtype}

\usepackage{graphicx}
\usepackage{amsmath}
\usepackage{amsfonts}

\usepackage{bbm}
\usepackage{enumitem}

\usepackage{multirow}

\newcommand{\zhenjie}[1]{{#1}}
\newcommand{\camera}[1]{{#1}}
\title{Educational Question Generation of Children Storybooks via Question Type Distribution Learning and Event-Centric Summarization}

\author{\normalsize{Zhenjie Zhao$^{\spadesuit~\heartsuit}$~~Yufang Hou$^{\clubsuit}$~~Dakuo Wang$^{\diamondsuit}$~~Mo Yu$^{\bigstar}$~~Chengzhong Liu$^{\blacklozenge}$~~Xiaojuan Ma$^{\blacklozenge}$}\\
 \normalsize{$^{\spadesuit}$~Nanjing University of Information Science and Technology}~~~  \normalsize{$^{\heartsuit}$~Nankai University}  \\
  \normalsize{$^{\clubsuit}$~IBM Research Europe~~~$^{\diamondsuit}$~IBM Research}~~~
  \normalsize{$^{\bigstar}$~WeChat AI} \\
  \normalsize{$^{\blacklozenge}$~The Hong Kong University of Science and Technology} \\
  \normalsize{\tt zzhaoao@nuist.edu.cn}, {\tt yhou@ie.ibm.com}, {\tt dakuo.wang@ibm.com} \\ \normalsize{\tt moyumyu@tencent.com, chengzhong.liu@connect.ust.hk, mxj@cse.ust.hk} \\
  }

\begin{document}
\maketitle
\begin{abstract}
Generating educational questions of fairytales or storybooks is vital for
improving children's literacy ability. 
However, it is challenging to generate questions that capture the interesting aspects of a fairytale story with educational meaningfulness. 
In this paper, we propose a novel question generation method
that first learns the question type distribution of an input story \camera{paragraph}, 
and then summarizes salient events 
which can be used to generate high-cognitive-demand questions.
To train the event-centric summarizer, we fine-tune a pre-trained transformer-based sequence-to-sequence model using silver samples composed by educational
question-answer pairs. On a newly proposed educational question-answering dataset \emph{FairytaleQA}, we show good performance of our method
on both automatic and human evaluation metrics. Our work indicates the necessity of decomposing question type distribution learning
and event-centric summary generation for educational question generation.
\end{abstract}

\input{sections/introduction}
\input{sections/relatedwork}
\input{sections/method}

\input{sections/experiment}

\input{sections/conclusion}
\bibliography{anthology,custom}
\bibliographystyle{acl_natbib}

\clearpage
\appendix

\section*{Appendix}

\input{sections/appendix}




\end{document}

%% file: sections/introduction.tex
\section{Introduction}
Listening to and understanding fairy tales or storybooks are very crucial for children's early intellectual and literacy development \cite{sim2014}.
During the storybook reading process, prompting suitable questions with educational purposes can help children understand the content and inspire their interests \cite{Zevenbergen2003DialogicRA,Ganotice2017EnhancingPR}. 

\zhenjie{There is an evidence that \textbf{high-cognitive-demand} (HCD) questions usually relate to 
good learning achievement \cite{doi:10.3102/00346543049001013}.
HCD questions usually correspond to application, analysis,synthesis, and evaluation questions in 
Bloom's taxonomyof cognitive process \cite{doi:10.3102/00346543049001013,Anderson2000ATF},
which are salient events merged from different elements across a session \cite{cd}.
}
However, it is challenging even for humans to ask educationally meaningful questions to engage children 
in storybook reading, which could be due to adults lacking the skills or time to integrate such interactive 
opportunities \cite{Golinkoff2019LanguageMD}.
Recent research shows that AI-powered conversational agents 
can play the role of language partners to read fairy tales to children and ask
them educational questions \cite{XU2021104059}. 
This motivates us to investigate techniques to generate \zhenjie{HCD} educational questions for children's storybooks automatically.
Automating the generation of such questions can have great value in supporting children's language development through guided conversation.%

    

During storybook reading, HCD questions require children to make inferences and predictions. In contrast to low-cognitive-demand \zhenjie{(LCD)} questions describing facts in stories (\textit{e.g.}, \emph{Who is Snow White's mother?}), HCD questions are often related to events and their relations (\textit{e.g.}, \emph{Why did the queen want to kill Snow White?} or \emph{What happened after the huntsman raised his dagger in the forest?}). 

Most previous work on question generation (QG) focuses on generating questions based on pre-defined answer spans 
\cite{krishna-iyyer-2019-generating,pyatkin-etal-2021-asking,cho-etal-2021-contrastive}. 
\zhenjie{Such systems that use ``keywords'' or specific events 
often generate LCD questions 
that are factual questions based on local context, 
but cannot work well on HCD cases, where we need to capture the
salient events and understand the relations across multiple elements/events. }
Recently, \citet{2109.03423} released a fairytale question answering dataset \textbf{FairytaleQA} 
containing around 10.5k question-answer pairs annotated by education experts. 
\zhenjie{Each question is assigned to a specific type, and some types,
such as ``\emph{action}'', ``\emph{causal relationship}'', are high-cognitive-demanding.}
This makes it possible to investigate generating educational questions to support children's interactive storybook reading.

In this paper, we propose a novel framework combining question type prediction and event-centric summarization to generate educational questions for storybooks.
In the first stage, we learn to predict the question type distribution for a given input 
and add pseudo-label so that after prediction, 
we can know both the types of questions and how many questions of each type. In the second stage, conditioned on 
question types and the order of the question under the current question type,
we extract salient events that are most likely for educators to design questions on
and then generate an event-centric summarization of the original input. 
Finally, in the third stage, we use the output of the second stage to generate questions. Each summarization is used to generate one question. 
Note that it is difficult to obtain gold annotations for event-centric summarization. Instead, we rewrite annotated questions, and their corresponding
hypothesized answers into question-answer statements \cite{1809.02922} as silver training samples. We hypothesize that HCD questions are around main plots in narratives and can guide our summarization model to focus on salient events.
We evaluate our system on the FairytaleQA dataset and show the superiority of the proposed method on both automatic and human evaluation metrics compared to strong baselines.


    
    


%% file: sections/relatedwork.tex
\section{Related Work}

\subsection{Question Generation}
Question answering based on context
has achieved remarkable results \cite{rajpurkar-etal-2016-squad,Zhang2020MachineRC}. 
The reverse problem, namely, question generation \cite{duan-etal-2017-question,chan-fan-2019-recurrent}, usually relies on pre-selecting spans from an input text as answers and a single sentence as the context. 
However, to generate questions across a long paragraph 
 in which the key information may come from  multiple different sentences in fairy tales \cite{2109.03423}, these existing models relying on one text segment usually do not work well. 

A few studies are focusing on generating questions that are based on multi-sentence or multi-document information fusion
\cite{pan-etal-2020-semantic,xie-etal-2020-exploring,Tuan_Shah_Barzilay_2020}. 
NarrativeQA \cite{kocisky-etal-2018-narrativeqa} is an effort that tries to integrate key information across multiple locations of a paragraph for question answering/generation.
Similarly, MS MARCO \cite{nguyen2016ms} is a dataset that integrates multiple locations of answers for users' queries
in search engines. In \newcite{cho-etal-2021-contrastive}, a contrastive method is proposed that first trains a supervised model to generate questions based on a single document and then uses a reinforcement learning agent 
to align multiple questions from multiple documents. In \newcite{lyu-etal-2021-improving}, the authors use a rule-based method to generate questions with summaries and report to achieve good performance. 

The methods mentioned above usually do not consider the educational dimension and may not work well on fairy tales.
Considering our research focus of fairytales, it is vital to generate questions that have educational purposes.
In FairytaleQA \cite{2109.03423}, experts usually write different types of questions for separate paragraphs. We hypothesize that context plays a significant role in deciding the type of questions that should be asked during the interactive storybook reading with children.
Therefore 
it is necessary to investigate not only how to summarize salient events but also how to learn the question type distribution. 


\subsection{Text Summarization}
Summarization methods can be classified into extractive summarization and abstractive summarization.
Extractive methods select sentences from the source documents to compose a summary; abstractive methods applies neural generative models to
generate the summary token-by-token. 

Extractive summarization methods, such as TextRank \cite{mihalcea-tarau-2004-textrank}, feature-based methods 
\cite{jagadeesh2005sentence,5392672,10.5555/3298483.3298681}, and topic-based methods \cite{ozsoy-etal-2010-text}, 
do not work to generate HCD questions on the fairytale scenario because such questions often are based on multiple sentences. 

Abstractive methods based on encoder-decoder architectures usually encode an input document
token-by-token sequentially \cite{rush-etal-2015-neural} and cannot capture the fine-grained hierarchical relations in a document,
such as actions, causal relationships. Graph neural network (GNN) models are recently used in summarization research 
\cite{2106.06090,wang-etal-2020-heterogeneous,xu-etal-2020-discourse,li-etal-2021-timeline}, thanks to their 
ability to model the complex relations in a document. 
For example, in \newcite{xu-etal-2020-discourse}, researchers used a discourse-level dependency
graph to encode a document and then decoded discourse-level embeddings to select sentences extractively.
Similarly, in \newcite{wang-etal-2020-heterogeneous}, researchers have used a heterogeneous graph to encode both 
token-level and sentence-level relations in a document and then used it to extract sentences.
Still, in the education domain,
summarizing salient events of one paragraph that 
can be used to generate educational questions is an open problem. 
In this paper, 
we develop an event-centric summarization method based on BART \cite{lewis-etal-2020-bart}. To obtain the training data, we compose 
educational question-answer pairs through a rule-based method and use them as silver ground-truth samples.


%% file: sections/method.tex
\section{Method}

\begin{figure*}
    \centering
    \includegraphics[width=0.7\textwidth]{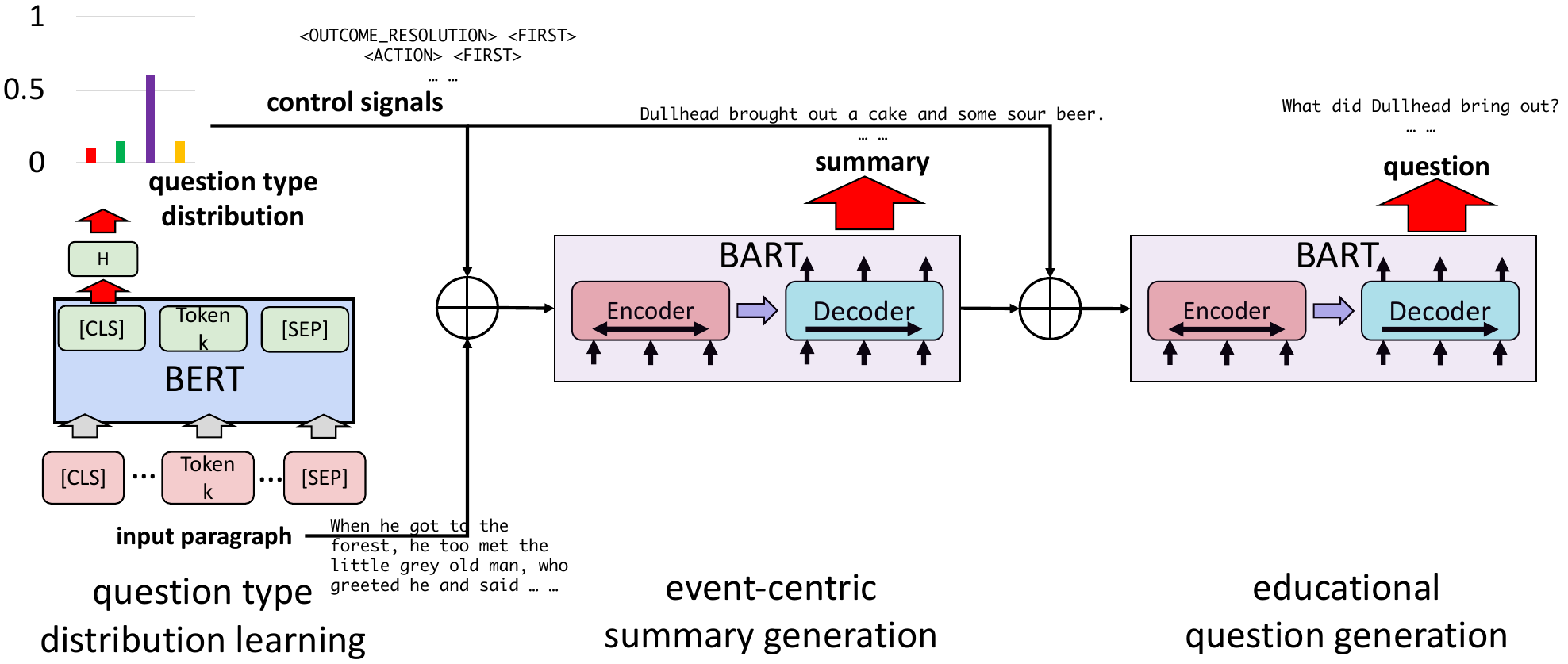}
    \vspace{-5pt}
    \caption{The overview of our educational question generation system of fairy tales.}
    \label{overview}
\end{figure*}

The overview of our educational question generation system 
for storybooks
is shown in Figure \ref{overview},
which contains three modules: question type distribution learning, event-centric summary generation,
and educational question generation.

Given an input paragraph $d$, we first predict the type distribution of output questions $\boldsymbol{p} = (p_1, p_2, \ldots, p_T)$,
where $p_i$ denotes the probability of question type $i$, $T$ is the total number of question types. We then
transform the distribution into the number of questions under each question type $\boldsymbol{l} = (l_1, l_2, \ldots, l_T)$.
Afterwards, we first generate $l_i$ summaries of type $i$ with the input paragraph $d$, and then 
generate $l_i$ questions of type $i$ with the corresponding summaries. 

\subsection{Question Type Distribution Learning}

We fine-tuned a BERT model \cite{devlin-etal-2019-bert}, and adapt the output $m$ dimensional class token $\boldsymbol{h}_c \in \mathbb{R}^m$ to learn the question type distribution. 
Specifically, the predicted distribution is obtained by
$p_i = \frac{e^{(\boldsymbol{W}\boldsymbol{h}_c+\boldsymbol{b})_i}}{\sum_{i=1}^T e^{(\boldsymbol{W}\boldsymbol{h}_c+\boldsymbol{b})_i}}$, where $\boldsymbol{W} \in \mathbb{R}^{T\times m}, \boldsymbol{b} \in \mathbb{R}^T$ are learnable parameters, 
$(\cdot)_i$ denotes the operator of selecting the $i$-th element of a vector.

Assuming there are $N$ training samples, we minimize the K-L divergence loss $\mathcal{L}_{K-L} = \sum_{j=1}^N \frac{1}{N} \sum_{i=1}^{T}p_i^{(j)} \log \frac{p_i^{(j)}}{\hat{p}_i^{(j)}}$, where $p_i^{(j)}$ denotes the probability of question type $i$ for the $j$-th sample, and $\hat{p}_i^{(j)}$ is our predicted value.

To improve the prediction performance, similar to \newcite{ijcai2018-639}, we also conduct a multi-label classification task,
where we use the question type with the maximal probability as the class of the output.
In particular, we add a cross entropy loss $\mathcal{L}_{CE} = -\sum_{j=1}^N \frac{1}{N} \sum_{i=1}^T \mathbbm{1}(y_i^{(j)}) \log \hat{y}_i^{(j)}$, where $\mathbbm{1}(y_i^{(j)})$ equals to $1$ if $i$ is the question type with the maximal probability for the sample $j$.

In summary, we conduct a multi-task learning for question type distribution prediction, and the final training loss is a weighted sum of the K-L loss and the cross entropy loss: $\mathcal{L} = \gamma \mathcal{L}_{K-L} + (1-\gamma) \mathcal{L}_{CE}$, where $\gamma$ is a weight factor.

To predict the number of questions for each question type during training, we add a pseudo label $1$ to the
original label $\boldsymbol{l}=(l_1, l_2, \ldots, l_n)$, \textit{i.e.}, $\boldsymbol{l}=(l_1, l_2, \ldots, l_n, 1)$.
We can then normalize it to get the ground-truth probability distribution $\small{\boldsymbol{l}=(\frac{l_1}{\sum_{k=1}^n l_k +1}, \ldots, \frac{l_n}{\sum_{k=1}^n l_k +1}, \frac{1}{\sum_{k=1}^n l_k +1})}$. During testing, assuming we get the predicted
distribution $\boldsymbol{p}=(p_1, p_2, \ldots, p_n, p_{pseudo})$, 
we can obtain the number of each type of 
questions by diving the probability of this pseudo label $p_{pseudo}$ as: $n_i = \lfloor \frac{p_i}{p_{pseudo}} + 0.5\rfloor$.

\subsection{Event-centric Summary Generation}

In FairytaleQA, one paragraph usually has multiple questions with different question types, and information in one educational question may 
scatter across multiple parts. 
As mentioned before, we assume that context
plays a big role to decide the type and the number of questions to 
be asked during the interactive storybook reading,
and HCD questions are around salient events and the relations. 
With the output from the previous component, 
we can use the predicted question type distribution as a control signal, and select corresponding events for one particular question type.

In particular, we add two control signals before an input paragraph: question type signal \texttt{<t>} and question order signal \texttt{<c>}, where \texttt{<t>} $\in$ \texttt{T}, \texttt{<c>} $\in$ \texttt{C},
\texttt{T} denotes the set of all question types, \texttt{C} denotes the set of order, \textit{i.e.}, \{\texttt{<first>, <second>, <third>,} ...\}. We train a BART summarization model \cite{lewis-etal-2020-bart} to conduct the event-centric summary generation task. The input of the BART model is: \texttt{<t> <c> d}, and the output of the BART model is a summary that collects related events for an educational question type, where \texttt{d} denotes the input paragraph.

Obtaining the golden summaries is difficult.
However, a QA dataset, like FairytaleQA, provides both questions and their corresponding answers.
We can therefore re-write the annotated questions and answers together to obtain question-answer statements, 
which are used as silver summaries to train our summarization model. 
We used the rule-based method in \newcite{1809.02922} which inserts answers into the semantic parsed questions and eliminates question words.


\subsection{Educational Question Generation}
With the summary generated in the second stage, generating an educational question is fairly straightforward. Because the summary has already contained all key events for the target educational question type, we can train a question generation model directly on top of it using the annotated questions. 
We fine-tune another BART model to generate questions, with the type and order control signals added before the input summary to control the generated results. Note that our question generation model does not reply on pre-selected answer spans.



%% file: sections/experiment.tex
\section{Experimental Setup}

To demonstrate the effectiveness of our proposed method, we conducted a set of 
experiments on the FairytaleQA dataset. 

\subsection{Dataset}
The FairytaleQA dataset \cite{2109.03423} contains annotations of 278 books, including 232 training books, 23 test books, and 23 validation books. 
Each book has multiple paragraphs, and for each paragraph of one book, 
there are several educational question-answer pairs annotated by education experts. 
\zhenjie{The question type distribution is consistent among annotators.}
In total, there are seven types:

\noindent$\bullet$\textbf{Character}: questions that contain the character of the story as the subject and ask for additional information about that character;

\noindent$\bullet$\textbf{Setting}: questions that start with ``Where/When'';

\noindent$\bullet$\textbf{Feeling}: questions that start with ``How did/do/does X feel?'';

\noindent$\bullet$\textbf{Action}: questions that start with "What did/do/does X do?" or "How did/do/does X" \textit{or} questions that contain a focal action and ask for additional information about that action;

\noindent$\bullet$\textbf{Causal relationship}: questions that start with ``Why'' or ``What made/make'';

\noindent$\bullet$\textbf{Outcome resolution}: questions ask about logic relations between two events, such as ``What happened...after...'';

\noindent$\bullet$\textbf{Prediction}: questions that start with ``What will/would happen...''.

\zhenjie{The first three
are factual questions that are low-cognitive-demanding},
and can be handled well by traditional span-based question generation methods \cite{2109.03423}.
\zhenjie{The remaining four types
usually require people to make inference from multiple elements \cite{PARIS}, which 
correspond to high-level cognitive skills in Bloom's taxonomy \cite{Anderson2000ATF}, and 
can be viewed as HCD questions.}
For the question type \textit{prediction}, it usually asks for events that do not appear in storybooks, which is not our focus in this paper.
We only consider \textit{action, causal relationship}, and \textit{outcome resolution}. 
\zhenjie{There is \camera{a small portion (985 out of 10580) of} questions that span multiple paragraphs. 
To control the cognitive-demand level for children, we also removed those questions.}
The statistics of the selected data is shown in \camera{section A of} the appendix.





\subsection{Baselines}

We compared our system with two baselines: 1) 
the method proposed in \newcite{2109.03423} (denoted as QAG), which is the only method that considers generating
educational questions; 
2) using FairytaleQA, we trained an end-to-end BART model.











\vspace{3mm}
\noindent \textbf{QAG.} 
\zhenjie{The QAG model \cite{2109.03423} use ``keywords'' (semantic role labelling) to
identify entities and events and then generate questions, which} contains four steps: 
1) generate a set of answers based on semantic roles of verbs;
2) generate questions based on these answers; 3) generate answers based on the questions generated in the second step; 
4) rank generated question-answer pairs and choose the top questions.  
We trained the question generation model in the second step and the answer generation model in the third step using the selected questions.
We use the top 10/5/3/2/1 generated questions as baselines, denoted as QAG (top10), QAG (top5), QAG (top3), QAG (top2), and QAG (top1), respectively.

\vspace{3mm}
\noindent \textbf{E2E.} Using FairytaleQA with question types \textit{action, causal relationship}, and \textit{outcome resolution},
we trained one BART-large model to generate questions based one paragraph end-to-end. During testing, we used a maximal length $100$ tokens 
(roughly $7$ questions according to Table \ref{statistics2}) and selected the first $2$ questions as the output for evaluation.
We denote this method as E2E.

\begin{table*}[h]
    \centering
    \small
    \begin{tabular}{c|c|c|c|c|c|c}
    \hline \hline
    & \multicolumn{3}{c|}{Rouge-L} & \multicolumn{3}{|c}{BERTScore} \\ \hline
    Method & Pre (val/test) & Rec (val/test) & F1 (val/test) & Pre (val/test) & Rec (val/test) & F1 (val/test)\\ \hline
    E2E& 16.32/15.76& 36.21/35.89& 20.29/19.73 & 0.8855/0.8839& 0.8425/0.8407&0.8632/0.8615\\ \hline
    QAG (top1) & 34.58/32.33 & 19.56/19.69& 22.88/22.29& 0.8599/0.8623 & 0.8776/0.8770&0.8684/0.8694\\ \hline
    QAG (top2) & 28.45/26.58 & 30.51/30.34& 26.76/25.67& 0.8830/0.8810&0.8745/0.8702& 0.8786/0.8754\\ \hline
    QAG (top3) & 24.29/22.74 & 36.80/36.31& 26.67/25.50&0.8866/0.8846 & 0.8663/0.8629&0.8761/0.8734\\ \hline
    QAG (top5) & 20.38/19.25 & 43.45/43.04& 25.55/24.53& 0.8883/0.8862& 0.8571/0.8540&0.8722/0.8696\\ \hline
    QAG (top10)& 18.12/17.26 & \textbf{46.57/47.04}& 24.05/23.34& 0.8873/0.8848&0.8503/0.8472 &0.8681/0.8654\\ \hline
    Ours & \textbf{33.49}/\textbf{37.50} & 37.50/31.54 & \textbf{31.81}/\textbf{30.58}& \textbf{0.8915}/\textbf{0.8862}& \textbf{0.8886}/\textbf{0.8930}& \textbf{0.8898}/\textbf{0.8893}\\ \hline
    \end{tabular}
    \caption{The comparison results on Rouge-L and BERTScore by concatenating generated questions together.}
    \label{main_result}
\end{table*}

\subsection{Evaluation Metrics}

We adopt both automatic and human evaluation to measure the performance of our method. 

\subsubsection{Automatic Evaluation}

For automatic evaluation, similar to \newcite{2109.03423}, we use the Rouge-L score \cite{lin-2004-rouge}, and report the average precision, recall, and F1 values. Meanwhile, we also use BERTScore \cite{BERTScore} to evaluate the semantic similarity of generated questions with the ground-truth questions, and report the average precision, recall, and F1 values. 
\emph{In contrast to 
\newcite{2109.03423}, 
\camera{we mainly consider concatenating all generated questions into one sentence
and comparing it with the concatenated ground-truth questions.}} 
\zhenjie{This is because for each paragraph,
we need to evaluate the generated quality of not only each question but also the question type distribution 
for sub-skills required in education as a whole \cite{PARIS}.
Since the question order does not have much effects on Rouge-L, 
concatenating questions also partially takes individual question quality into consideration.
\camera{Moreover, we also consider the same setup used in \newcite{2109.03423} that
takes the max score of each gold question against the generated questions, then averages the scores of all generated questions.}}

\subsubsection{Human Evaluation}

To evaluate the quality of our generated questions and their educational significance, we further conducted a human evaluation
session. After regular group meetings, we concluded the following four dimensions, \zhenjie{
where children appropriateness is the  main metric for our educational application}:

\noindent 1. \textbf{Question type}: whether the generated questions belong to any of the three event types.

\noindent 2. \textbf{Validity}: whether the generated questions are valid questions according to the original paragraph.

\noindent 3. \textbf{Readability}: whether the generated questions are coherent and grammatically correct.


\noindent 4. \textbf{Children appropriateness}: \zhenjie{to what extent} would you like to ask this question when you read the story to a five year's old child?


\subsection{Implementation Details}

For re-writing silver summaries,
there are 8 sentences that cannot be parsed successfully. 
In this case, we wrote the silver statements manually. 
We also corrected 5 low-quality statements manually.

The weight factor for question type distribution learning is set as $0.7$ empirically.
For question type distribution learning, we used a BERT cased large model.
For summary generation, we used a BART cased base model. 
For question generation, we used a BART cased large model.
The batch sizes of all training are set as $1$. For the generation process, we only used a greedy decoding method. 
\camera{Automatic evaluation results were calculated with open sourced packages 
\footnote{\camera{We used the package from \url{https://github.com/google-research/google-research/tree/master/rouge} to calculate Rouge-L,
and the package from \url{https://github.com/Tiiiger/bert_score}} to calculate BERTScore.}.} 
For all methods, we removed duplicated questions and questions that has less than 3 tokens.
All experiments were conducted on a Ubuntu server with Intel(R) Xeon(R) Silver 4216 CPU @ 2.10GHz,
32G Memory, Nvidia GPU 2080Ti, Ubuntu 16.04. \zhenjie{Training our model took about three hours.}

\section{Results and Analysis}

\subsection{Automatic Evaluation Results}

\begin{table}[t]
    \centering
    \small
    \begin{tabular}{|c|c|c|c|}
    \hline
    Method & Pre(val/test) & Rec(val/test) & F1(val/test) \\ \hline
    E2E& 31.29/30.80 & 36.21/36.53 & 31.77/31.65 \\ \hline
    QAG (top2)& 35.17/33.51 & 35.33/33.83 & 34.21/32.64 \\ \hline
    Ours & \textbf{48.30}/\textbf{44.05} & \textbf{39.55}/\textbf{36.68} & \textbf{41.78}/\textbf{38.29}\\ \hline
    \end{tabular}
    \caption{\camera{The comparison results with the setup used by \newcite{2109.03423}.}}
    \label{arthur_evaluation}
\end{table} 

The results of automatic evaluation on both validation and test datasets are shown in Table \ref{main_result}. For Rouge-L,
compared to E2E and QAG, our method can achieve the best results except for the recall values. In particular, our method outperforms E2E by about 20 points, and outperforms 
the best QAG model (top2) by about 10 points on the precision scores. For F1, our method outperforms E2E by about 10 points, and outperforms the best QAG model (top2) by about 5 points. These results show that our method can match the ground-truth questions lexically better than other methods. However, the recall score of our method is not as good as E2E and QAG (top5 \& 10). This is because for E2E and QAG (top5 \& 10), they generally generate more questions than our method\footnote{On the test data, the mean of the generated questions by our method is 1.9 (std: 0.6), which is closer to the case of ground-truth (mean: 2.2, std: 1.5)}.
For BERTScore, our method achieves the best results on precision, recall, and F1.
\zhenjie{Although our method outperforms QAG (top2) by a small margin, it still
outperforms other QAG models by at least 1 point.}
\camera{For the setup used by \newcite{2109.03423}, as shown in Table \ref{arthur_evaluation},
our method also outperforms the best QAG model, \textit{i.e.}, QAG (top 2), and E2E by a
large margin in terms of Rouge-L. 
We believe that decomposing question types explicitly and using event-centric summaries to generate questions 
can capture the internal structure of educational question annotations and fit the data distribution in a more 
accurate way.
}

\zhenjie{Some examples of the generated questions can be seen in Table \ref{generated}.
Our method usually can predict the correct question types, and cross multiple elements to generate HCD questions, with a limitation of factuality errors. 
More examples and comparison can be found in \camera{section C} of the appendix.}

\begin{table*}[t]
    \centering
    \small
    \begin{tabular}{|l|llll|}
    \hline
                          & \multicolumn{4}{p{14cm}|}{Questions}                                                                                                                            \\ \hline
    \multirow{2}{*}{\hfil  QAG (top2)}       & \multicolumn{4}{p{14cm}|}{\textbf{P1}: \camera{Once upon a time there was a farmer who had carted pears to market .? Why did the farmer want to cart pears?}}                 \\ \cline{2-5} 
    & \multicolumn{4}{p{14cm}|}{\textbf{P2}: \camera{What happened to the dwarf after he left? As for the silent earl and his irish sweetheart , they were married as soon?}}   \\ \hline  
    \multirow{2}{*}{\hfil Ours}       & \multicolumn{4}{p{14cm}|}{\textbf{P1}: Why did the bonze want to get a good price for the pears? (causal relationship) What did the bonze ask for? (action)}                 \\ \cline{2-5} 
                                & \multicolumn{4}{p{14cm}|}{\textbf{P2}: What did the Islanders want to express when they were married? (action) Why did the Islanders hold to the belief that Snorro was spirited away? (causal relationship)}                                                                                                                                     \\ \hline
    \multirow{2}{*}{\hfil Gold}       & \multicolumn{4}{p{14cm}|}{\textbf{P1}: Why did the farmer hope to get a good price for the pears? (causal relationship) What did the farmer do when he grew angry? (action)} \\ \cline{2-5} 
                                & \multicolumn{4}{p{14cm}|}{\textbf{P2}: What did Paul and Lady Morna do after Harold's funeral was over? (action) Why did Snorro lose all chance of finding the magic carbuncle? (causal relationship)}                                                                                                                                     \\ \hline
    \end{tabular}
    \caption{\zhenjie{\camera{Randomly selected examples} of generated questions from two paragraphs (\textbf{P1} and \textbf{P2}).}}
    \label{generated}
\end{table*}



Apart from the overall performance, we also investigated the performance of each module of our method. Because the performance values on both the validation and test data are similar, to simplify our experiment, in the following sections, we only conducted experiments on the test data.


\paragraph{Question Type Distribution Learning.}


On the test set, the K-L divergence between the prediction results of our BERT-based model and 
ground-truth is $0.0089$, which shows that the performance of our question type distribution learning module is relatively satisfactory.
We also use the ground-truth question type distribution as an input and calculate the final Rouge-L score with our system.
The results are shown in Table \ref{gt_tdl}. Compared to the ground-truth question type distribution, our system still has 
lower precision and F1 scores. Having a more accurate question type distribution prediction is beneficial for improving the 
overall performance.

\begin{table}[t]
    \centering
    \small
    \begin{tabular}{|c|c|c|c|}
    \hline
    Method & Pre & Rec & F1 \\ \hline
    Ours (gt)& \textbf{46.48}& \textbf{31.96} & \textbf{35.77} \\ \hline
    Ours (tdl) & 37.50& 31.54 & 30.58\\ \hline
    \end{tabular}
    \caption{The Rouge-L scores of our method with the ground-truth (denoted as gt) and predicted (denoted as tdl) on question type distribution learning.}
    \label{gt_tdl}
\end{table}

\paragraph{Event-centric Summary Generation}

To investigate the quality of the generated summaries, we compare the generated results with the silver summary 
ground-truth. Similar to the evaluation method of generated questions, 
we concatenated the generated summaries and calculated the Rouge-L score with the concatenated ground-truth summaries.
The results are $15.41$ precision, $30.60$ recall, and $18.85$ F1, 
which shows that there is still a lot of room to improve the summarization module.

\paragraph{Upper-bound Results with Silver Summary}
To see how the upper-bound performance is if we have perfect summaries,
we input the silver summaries to our educational 
question generation model. 
The Rouge-L scores of generated questions are $92.71$ precision, $85.65$ recall, $87.67$ F1,
which shows the potential that once a good summary containing salient events is available, generating an educational question
is relatively easy. The core challenge is to obtain good summaries, which we believe will be a valuable next step in future work. 

\subsection{Human Evaluation Results}
We conducted a human evaluation \zhenjie{with consent} of our method against the best-performed baseline QAG (top2).
We first randomly sampled 10 books from the test set. For each book, we randomly sampled 5 paragraphs. We then conducted experiments to evaluate the generated results on question type and quality.
\zhenjie{Participants are researchers or PhD students based in Europe, U.S., and China working
on natural language processing and human-computer interaction in the education domain with 
at least 3 years of experience, and were recruited through word-of-mouth and paid \$30. We had a training session to ensure the annotation
among participants is consistent. This study is approved by IRB.}

\paragraph{Question type.}
Three human participants annotated the types of all generated questions. The inter-coder reliability score (Krippendoff's alpha \cite{krippendorff2011computing}) among three participants is 0.86,
indicating a relatively high consistency. The annotated results are shown in Table \ref{human_evaluation1}. 
Overall, our method demonstrates a much smaller K-L distance (\textbf{0.28}) to the ground-truth distribution, compared to QAG (\textbf{0.60}).
We can see that our method has a better estimation of the distribution of question types, which is closer to the distribution of the ground-truth. QAG has a biased question type distribution and generates more outcome resolution questions.

\begin{table}[h]
    \small
    \centering
    \begin{tabular}{|c|c|c|c|}
    \hline
    & QAG(top2) & Ours & Ground-truth\\ \hline
    Vague & 17/17\% & 15/17\% & 0/0\% \\ \hline
    Action & 21/21\% & 34/38\% & 47/48\% \\ \hline
    Causal & 10/10\% & 36/40\% & 32/33\% \\ \hline
    Outcome & 51/52\% & 5/6\% & 18/19\% \\ \hline
    \end{tabular}
    \caption{The human evaluation results on of question types (\textbf{vague} denotes question types that are hard to decide or questions that have grammar mistakes).}
    \label{human_evaluation1}
\end{table}

\paragraph{Question quality.}

We invited another five human participants and conducted a human evaluation to further evaluate the quality of the generated questions from our model against the ground-truth and QAG, including \emph{validity}, \emph{readability}, and \emph{children appropriateness}. 
Among the three dimensions, the \emph{children appropriateness} is most closely related to the educational purpose; the former two dimensions mainly measure the factual correctness and fluency  respectively.

For the total $10\times5$ paragraphs, each participant is assigned $20$ different paragraphs randomly, and each paragraph has annotation results from two participants.
For each paragraph, participants need to read the paragraph and its corresponding
questions and answers, and then rate the three dimensions on a five-point Likert-scale. 
The Krippendoff's alpha scores along the four dimensions are between 0.60 and 0.80 (validity: 0.80, readability: 0.69, 
children appropriateness: 0.60), indicating an acceptable consistency 
\cite{gretz-etal-2020-workweek}.

We conducted an independent-samples t-test to compare the performance of each model. 
Our model is significantly better than QAG on the main evaluation dimension of \emph{children appropriateness}: the mean score of our model and QAG are 2.56 and 2.22, with corresponding standard derivation 1.31 and 1.20 respectively.
This gives a significant score with p-value$=$0.009, showing that the questions generated by our model can indeed better fit the education scenario.
For reference, the ground-truth has a mean score and standard derivation of 3.96 and 1.02, indicating a still large space to improve.

On \emph{validity} and \emph{readability}, our model is on par with QAG. This is not surprising because both models are based on large pre-trained BART models that are good at generating natural and fluent sentences.
For validity, our model (avg: 3.19, std: 1.53) is a bit lower than QAG (avg: 3.27, std: 1.62);
for readability, our model (avg: 4.19, std: 1.53) is a bit higher than QAG (avg: 4.12, std: 1.33). 
A further breakdown in Table~\ref{human_evaluation2} shows that QAG wins mainly on action questions, because it directly generates questions conditioned on verbs. 
For causal relationship and outcome resolution questions, our method generally outperforms QAG. 




\begin{table}[h]
    \small
    \centering
    \begin{tabular}{|c|c|c|c|c|}
    \hline
    & QAG & Ours \\ \hline
    Vague & 2.06$^{*}$/2.03$^{*}$ & \textbf{2.97}$^{*}$/\textbf{3.03}$^{*}$ \\ \hline
    Action & \textbf{3.69}/\textbf{4.76}$^*$ & 3.35/4.34$^*$ \\ \hline
    Causal & \textbf{3.45}/4.45 & 3.10/\textbf{4.46} \\ \hline
    Outcome & 3.46/4.49$^{\*}$ & \textbf{3.50}/\textbf{4.80}$^{\*}$ \\ \hline
    \end{tabular}
    \caption{The mean values of human evaluation on question qualities (validity/readability), where $^*$ denotes significant difference.}
    \label{human_evaluation2}
\end{table}


\section{System Analysis}

To further investigate the effectiveness of our method, we conducted a set of ablation studies.

\subsection{Question Type Distribution Learning}

To investigate the effects of our question type distribution learning, we conducted a comparison study. 
In particular, we removed the question type distribution learning module (denoted as w/o tdl), and directly trained the summarization
and question generation models. In other words, during training, we concatenate all silver summaries as the output of the summarization model. During testing, we extract the first $2$ sentences as the predicted summaries. 
The results are shown in Table \ref{qtdl}. From the comparison, we can see that 
without knowing question types, the Rouge-L scores drop \zhenjie{about 3 points overall}, which implies the importance of our question 
type distribution learning module.

\begin{table}[t]
    \centering
    \small
    \begin{tabular}{|c|c|c|c|}
    \hline
    Method & Pre & Rec & F1 \\ \hline
    Ours (w/o tdl)& 32.62& 29.89 & 27.42 \\ \hline
    Ours & \textbf{37.50}& \textbf{31.54} & \textbf{30.58}\\ \hline
    \end{tabular}
    \caption{The Rouge-L scores of our method with and without question type distribution learning.}
    \label{qtdl}
\end{table}

\subsection{Event-centric Summary Generation}

To investigate the effects of our event-centric summary generation module, we conducted a comparison with different summarization
methods. The summarization methods include:
\textbf{1) Lead3}. We select the first three sentences of a \camera{paragraph} as the summary, and use them as input to the question generation model;
\textbf{2) Last3}. We select the last three sentences of a \camera{paragraph} as the summary, and use them as input to the question generation model.
\textbf{3) Random3}. We select the random three sentences of a \camera{paragraph} as the summary, and use them as input to the question generation model.
\textbf{4) Total}. We use each sentence of a \camera{paragraph} as the summary, and use them as input to the question generation model.
\textbf{5) TextRank}. TextRank is a typical extractive summarization method. We use TextRank to extract a summary, and for each sentence 
in the summary, we input it to the question generation model.

For other summarization methods, they cannot get the question type distribution like our method. For a fair comparison, 
we also remove the question type distribution learning module of our method, which is the same as the setting in section 6.1.
The results are shown in Table \ref{summary}, from which we can see that
extracting sentences from the paragraph is not enough for covering salient events for educational question generation.
Our event-centric summary generation method is an effective way for extracting educational events of fairy tales. Using all sentences (total) can have the highest recall score at the expense of accuracy, but the overall F1 score is still relatively low.

\begin{table}[t]
    \centering
    \small
    \begin{tabular}{|c|c|c|c|}
    \hline
    Method & Pre & Rec & F1 \\ \hline
    Lead3  & 25.20 &30.76& 24.73  \\ \hline
    Last3 & 24.35& 29.97 & 24.05\\ \hline
    Random3  & 23.75&28.88&23.07 \\ \hline
    Total  &22.69 &\textbf{34.34}&24.63\\ \hline
    TextRank  & 30.72&21.74& 21.94\\ \hline
    Ours (w/o tdl)& \textbf{32.62}& 29.89 & \textbf{27.42} \\ \hline
    \end{tabular}
    \caption{The comparison results (Rouge-L on question generation) of different summarization methods.}
    \label{summary}
\end{table}
















\begin{table}[t]
    \centering
    \small
    \begin{tabular}{|c|c|c|c|}
    \hline
    Method & Pre & Rec & F1 \\ \hline
    Action& 35.97&20.68& 24.29\\ \hline
    Causal& 13.70&11.23& 11.54\\ \hline
    Outcome& 6.15&4.97& 5.30\\ \hline
    Ours (individual) & 25.71& \textbf{33.08} & 26.27\\ \hline
    Ours (overall) & \textbf{37.50}& 31.54 & \textbf{30.58}\\ \hline
    \end{tabular}
    \caption{The comparison results of training separate summarization and question generation models on each question type.}
    \label{type}
\end{table}

\subsection{Multi-task Learning of Question Types}

Currently, we use control signals to constrain generating questions of different types, which can be viewed as a multi-task learning framework for multi-type question generation.
To investigate whether sharing parameters is a good way 
for our task, we trained individual summarization and question generation models using different question types. The results in Rouge-L are shown in 
Table \ref{type}. We can find that sharing parameters generally can achieve better performance because of the use of more training data. For only using one type of training data, owing to the error of question type distribution learning, the performance drops a lot, showing the importance of combining question type distribution learning and multi-task learning with different types of training data.

%% file: sections/conclusion.tex
\section{Conclusion}
In this paper, we propose a novel method for educational question generation for fairy tales,
which can potentially be used in early childhood education.
Our method contains three modules: question type distribution learning, 
event-centric summary generation, and educational question generation.
Through question type distribution learning, we can decompose the challenges of 
educational question generation by extracting related events of one question type and 
generating educational questions with a short event-centric summary, which 
improves the performance significantly. On both automatic evaluation and human evaluation,
we show the potential of our method. In the future, we plan to further
investigate the event-centric summary generation module by considering discourse-level information
to improve the summarization performance \zhenjie{and improve the factuality error problem}. 
We are also interested in deploying the system in real scenarios to benefit childcare-related domains.

\section*{Acknowledgments}

The authors thank constructive suggestions from all
anonymous reviewers, as well as all participants in our human evaluation session. This work is supported by the Hong Kong General Research Fund (GRF) with grant No. 16203421. Zhenjie Zhao is supported by the National Natural Science
of China Foundation under Grant No. 62106109 and the Startup Foundation for Introducing Talent of
NUIST.

%% file: sections/appendix.tex
\section{Dataset Statistics}

\begin{figure}[h]
    \centering
    \includegraphics[width=0.45\textwidth]{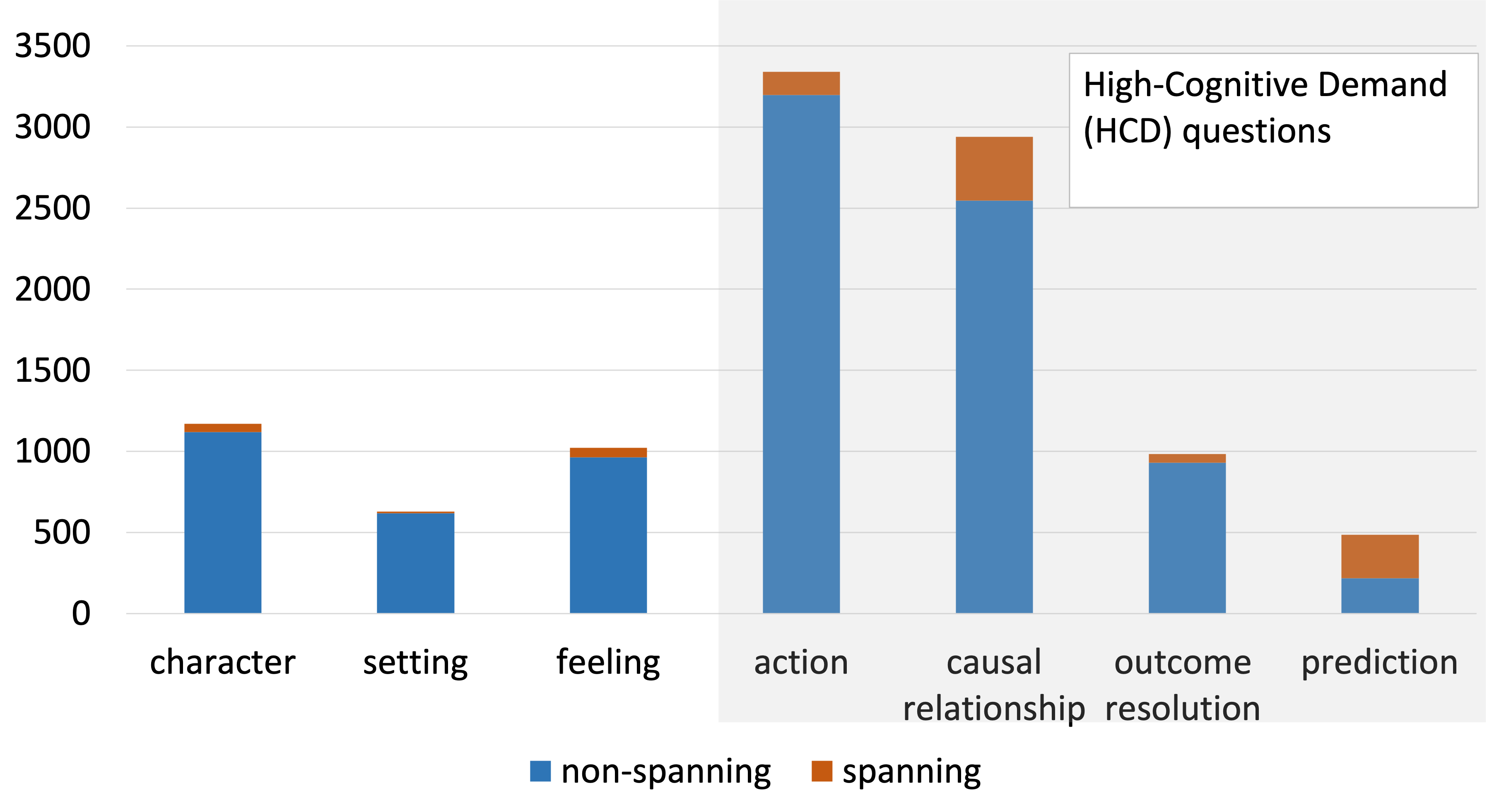}
    \caption{\camera{Question type distribution of FairytaleQA.}}
    \label{distribution}
\end{figure}

\camera{
The question type distribution of FairytaleQA is shown in Figure \ref{distribution}.
In particular, question types \textit{action, causal relationship, outcome resolution, and 
prediction} are considered as HCD questions.
For each question type, there are some questions that span multiple paragraphs 
(denoted as \textit{spanning}, \textit{character}: 53,
\textit{setting}: 11, \textit{feeling}: 59, 
\textit{action}: 143, \textit{causal relationship}: 392,
\textit{outcome resolution}: 54, \textit{prediction}: 266), which are discarded.
We select question types \textit{action, causal relationship, and outcome resolution} from FairytaleQA for 
conducting our experiments.
In total, there are 2998 out of 4095 paragraphs used,
including 2430 out of 3350 training paragraphs, 290 out of 380 validation paragraphs,
and 278 out of 365 paragraphs. 
The number of QA pairs for each question type and the total number 
are shown in Table \ref{statistics},
and the token-level statistics of the selected training data  
can be found in  Table \ref{statistics2}.}


\begin{table}[h]
    \centering
    \small
    \begin{tabular}{|c|c|c|c|c|}
    \hline 
    & train & val &test  & total \\ \hline
    \#action      &  2574  & 322 & 302 & 3198 \\  \hline
    \#causal relationship      & 2057 & 246 & 244 & 2547\\  \hline
    \#outcome resolution & 766 & 93 & 72 & 931\\ \hline
    \#selected  & 5397& 661& 618 & 6676\\  \hline
    \#total & 8548  & 1025 & 1007 & 10580 \\ \hline
    \end{tabular}
    \caption{\camera{The numbers of QA pairs for question types
    \textit{action}(\#action), \textit{causal relationship}(\#causal relationship) 
    and \textit{outcome resolution}(\#outcome resolution), the selected data (\#selected), 
    and all data of FairytaleQA (\#total).}}
    \label{statistics}
\end{table}

\begin{table}[h]
    \centering
    \small
    \begin{tabular}{|c|c|c|}
    \hline
    & mean & std \\ \hline
    \#question & 2.2 & 1.5 \\ \hline
    \#token (\camera{paragraph}) & 160.4 & 65.1 \\ \hline
    \#token (summary) & 17.8 & 7.2 \\ \hline
    \#token (question) & 10.1 & 3.1 \\ \hline
    \end{tabular}
    \caption{\camera{The mean and standard deviation (std) of the number of questions for each paragraph (\#question) 
    and the number of tokens (\#token) in paragraphs, summaries, 
    and questions in the training data.}}
    \label{statistics2}
\end{table}



\section{Potential Risks}

\camera{While High-Cognitive Demand (HCD) questions are considered in this paper, 
the cultivation of knowledge and ability is equally important for children.
The experiment results show that our method is competitive to generate HCD questions, and therefore 
it is helpful to improve children's cognitive ability. However, because of the unexplainability of end-to-end training,
we also find that sometimes our system may generate non-factual facts in terms of the original  context,
which has an potential risk on knowledge learning.
Owing to the factuality error problem of our system, 
we suggest to further investigate constructing structured knowledge of fairy tales and knowledge-grounded 
question generation for real-world applications.} 

\section{Examples of Generated Questions}

\zhenjie{Some \camera{randomly selected} examples of the generated questions can be found in Table \ref{example}.}



\begin{table}[h]
    \centering
    \small
    \begin{tabular}{p{7.5cm}}
    \hline \hline
    \vspace{1mm}

    \textbf{Paragraph}: Once upon a time there was a farmer who had carted pears to market. Since they were very sweet and fragrant, he hoped to get a good price for them. A bonze with a torn cap and tattered robe stepped up to his cart and asked for one. The farmer repulsed him, but the bonze did not go. Then the farmer grew angry and began to call him names. The bonze said: ``You have pears by the hundred in your cart. I only ask for one. Surely that does you no great injury. Why suddenly grow so angry about it?'' \\

    \vspace{1mm}

    \textbf{Gold questions}: Why did the farmer hope to get a good price for the pears? (causal relationship) What did the farmer do when he grew angry? (action) \\

    \vspace{1mm}

    \textbf{Generated questions by our method}: Why did the bonze want to get a good price for the pears? (causal relationship) What did the bonze ask for? (action) \\

    \vspace{1mm}

    \textbf{Generated questions by QAG (top2)}: Once upon a time there was a farmer who had carted pears to market .? Why did the farmer want to cart pears? \\

    \vspace{1mm}

    \textbf{Silver summaries}: The farmer hoped to get a good price for the pears because they were very sweet and fragrant. (causal relationship) The farmer called the bonze names when he grew angry. (action) \\

    \vspace{1mm}

    \textbf{Generated summaries by our method}: The bonze wanted to get a good price for the pears because they were very sweet and fragrant. (causal relationship) The bonze asked for one. (action) \\

    \hline \hline
    \end{tabular}
\end{table}

\begin{table}[H]
    \centering
    \small
    \begin{tabular}{p{7.5cm}}

    \hline \hline
    \textbf{Paragraph}: No one knew what had become of them; a few people were inclined to think that the Dwarf and his Raven had accompanied the Countess Fraukirk and the Countess Helga on their flight, but the greater part of the Islanders held to the belief, which I think was the true one, that the Powers of the Air spirited Snorro away, and shut him up in some unknown place as a punishment for his wickedness, 
    and that his Raven accompanied him.  At any rate, he was never seen again by any living person, and wherever he went, he lost all chance of finding the magic carbuncle. As for the Silent Earl and his Irish Sweetheart, they were married as soon as Earl Harold's funeral was over; and for hundreds of years afterwards, when the inhabitants of the Orkney Isles wanted to express great happiness, they said, ``As happy as Earl Paul and the Countess Morna.'' \\
    
    \vspace{1mm}

    \textbf{Gold questions}: What did Paul and Lady Morna do after Harold's funeral was over? (action) Why did Snorro lose all chance of finding the magic carbuncle? (causal relationship) \\

    \vspace{1mm}

    \textbf{Generated questions by our method}: What did the Islanders want to express when they were married? (action) Why did the Islanders hold to the belief that Snorro was spirited away? (causal relationship) \\

    \vspace{1mm}

    \textbf{Generated questions by QAG (top2)}: What happened to the dwarf after he left? As for the silent earl and his irish sweetheart , they were married as soon? \\

    \vspace{1mm}

    \textbf{Silver summaries}: Paul and Lady Morna they was married after Harold's funeral was over. (action) Snorro lost all chance of finding the magic carbuncle because he did not live on the island anymore. (causal relationship) \\

    \vspace{1mm}

    \textbf{Generated summaries by our method}: The Islanders wanted to express great happiness when they were married. (action) The Islanders held to the belief that Snorro was spirited away because he wickedness. (causal relationship) \\

\hline \hline
\vspace{1mm}

\textbf{Paragraph}: The King danced again with the beautiful maiden, and thought she had never looked so beautiful. And while he was dancing, he put a gold ring on her finger without her seeing it, and he commanded that the dance should last longer than usual. When it was finished he wanted to keep her hands in his, but she broke from him, and sprang so quickly away among the people that she vanished from his sight. She ran as fast as she could to her little room under the stairs, but because she had stayed too long beyond the half-hour, she could not stop to take off the beautiful dress, but only threw the fur cloak over it, and in her haste she did not make herself quite black with the soot, one finger remaining white. \\


    \textbf{Gold questions}: What did the king do to the maiden while they were dancing? (action) 
    What happened because the maiden stayed too long beyond the half-hour? (outcome resolution) 
    Why did the king want to keep the maiden's hand in his? (causal relationship) \\
    
    \vspace{1mm}
    
    \textbf{Generated questions by our method}: 
    What did the maiden do because it had stayed too long beyond the half-hour? (action) 
    What happened after the maiden danced again with her dress? (outcome resolution)
    Why did the maiden break from the King? (causal relationship) \\

    \vspace{1mm}

\textbf{Generated questions by QAG (top2)}: 
What happened after the king danced again with his beautiful maiden? 
What happened because the king danced again with his beautiful maiden?

\vspace{1mm}

\textbf{Silver summaries}: 
The king he put a gold ring on her finger without her seeing it to the maiden while they were dancing. (action) 
Because the maiden stayed too long beyond the half - hour she could not stop to take off the beautiful dress. (outcome resolution) 
The king wanted to keep the maiden's hand in his because he loved her. (causal relationship) \\

    \end{tabular}
\end{table}
    
\begin{table}[H]
    \centering
    \small
    \begin{tabular}{p{7.5cm}}


\textbf{Generated summaries by our method}: 
The maiden ran as fast as she could to her little room under the stairs because it had stayed too long beyond the half - hour. (action)
After the maiden danced again with her dress she only threw it over it, and in haste did not make herself quite black. (outcome resolution)
The maiden broke from the King because she had stayed too long beyond the half - hour. (causal relationship) \\

    \hline \hline
    \vspace{1mm}
    \textbf{Paragraph}:  
    Art thou satisfied now?' said Matte to his wife.  `I should be quite satisfied,' said his wife, 
    `if only I had two servants to help, and if I had some finer clothes. Don't you know that I am addressed as Madam?'  `Well, well,' said her husband. So Maie got several servants and clothes fit for a great lady.  `Everything would now be perfect if only we had a little better dwelling for summer. You might build us a two-storey house, and fetch soil to make a garden. Then you might make a little arbour up there to let us have a sea-view; and we might have a fiddler to fiddle to us of an evening, and a little steamer to take us to church in stormy weather.'  `Anything more?' asked Matte; but he did everything that his wife wished. The rock Ahtola became so grand and Maie so grand that all the sea-urchins and herring were lost in wonderment. Even Prince was fed on beefsteaks and cream scones till at last he was as round as a butter jar.  `Are you satisfied now?' asked Matte.  `I should be quite satisfied,' said Maie, `if only I had thirty cows. At least that number is required for such a household.'  `Go to the fairies,' said Matte.

    \vspace{1mm}

    \textbf{Gold questions}: What did Maie want Matte to build? (action) 
    How many cows did Maie want? (action) \\

    \vspace{1mm}

    \textbf{Generated questions by our method}: 
    What did Maie get? (action) 
    What did the rock Ahtola become? (action) \\

    \vspace{1mm}

    \textbf{Generated questions by QAG (top2)}: 
    What did matte ask his wife to do? 
    What did matte tell his wife to do?

    \vspace{1mm}

    \textbf{Silver summaries}: 
    Maie wanted a two - storey house Matte to build. (action) 
    Maie wanted thirty cows. (action) \\

    \vspace{1mm}

    \textbf{Generated summaries by our method}: 
    Maie got several servants and clothes fit for a great lady. (action)
    The rock Ahtola became so grand and Maie was lost in wonderment. (action) \\
    
    \hline \hline
    \end{tabular}
    \caption{\zhenjie{\camera{Randomly selected examples} of original paragraphs, their corresponding gold questions, questions generated
    by our method, questions generated by QAG (top2), silver summaries, and summaries generated by our method.}}
    \label{example}
\end{table}